\documentclass{article}
\usepackage{arXiv}

\usepackage{times}
\usepackage{xcolor}
\usepackage{soul}
\usepackage[utf8]{inputenc}
\usepackage[small]{caption}

\usepackage[pagebackref=false,breaklinks=true,letterpaper=true,colorlinks,bookmarks=false]{hyperref}
\usepackage{breakurl}
\usepackage{bm}
\usepackage{subfigure}
\usepackage{breakurl}
\usepackage{multirow}
\usepackage{epsfig}
\usepackage{graphicx}
\usepackage{amsmath}
\usepackage{amssymb}
\usepackage[titletoc]{appendix}
\usepackage{titlesec}

\usepackage{breakurl}

\title{SCAN: Sliding Convolutional Attention Network for Scene Text Recognition}

\author{
Yi-Chao Wu,
Fei Yin,
Xu-Yao Zhang,
Li Liu,
Cheng-Lin Liu,
\\
National Laboratory of Pattern Recognition (NLPR), Institute of Automation of Chinese Academy of Sciences \\
95 Zhongguancun East Road, Beijing 100190, Beijing, China \\
\{fyin,xyz,liucl\}@nlpr.ia.ac.cn,
}

\begin{document}

\maketitle

\begin{abstract}

% ¨¦?€¨¨???¡è¡ì?¡±????¨¨?¡­?¡°??¡°???€

  Scene text recognition has drawn great attentions in the community of computer vision and artificial intelligence due to its challenges and wide applications. State-of-the-art recurrent neural networks (RNN) based models map an input sequence to a variable length output sequence, but are usually applied in a black box manner and lack of transparency for further improvement, and the maintaining of the entire past hidden states prevents parallel computation in a sequence. In this paper, we investigate the intrinsic characteristics of text recognition, and inspired by human cognition mechanisms in reading texts, we propose a scene text recognition method with sliding convolutional attention network (SCAN). Similar to the eye movement during reading, the process of SCAN can be viewed as an alternation between saccades and visual fixations. Compared to the previous recurrent models, computations over all elements of SCAN can be fully parallelized during training. Experimental results on several challenging benchmarks, including the IIIT5k, SVT and ICDAR 2003/2013 datasets, demonstrate the superiority of SCAN over state-of-the-art methods in terms of both the model interpretability and performance.

\end{abstract}

\section{Introduction}

\begin{figure}[tb]
    \begin{center}
    \includegraphics[width=0.8\linewidth]{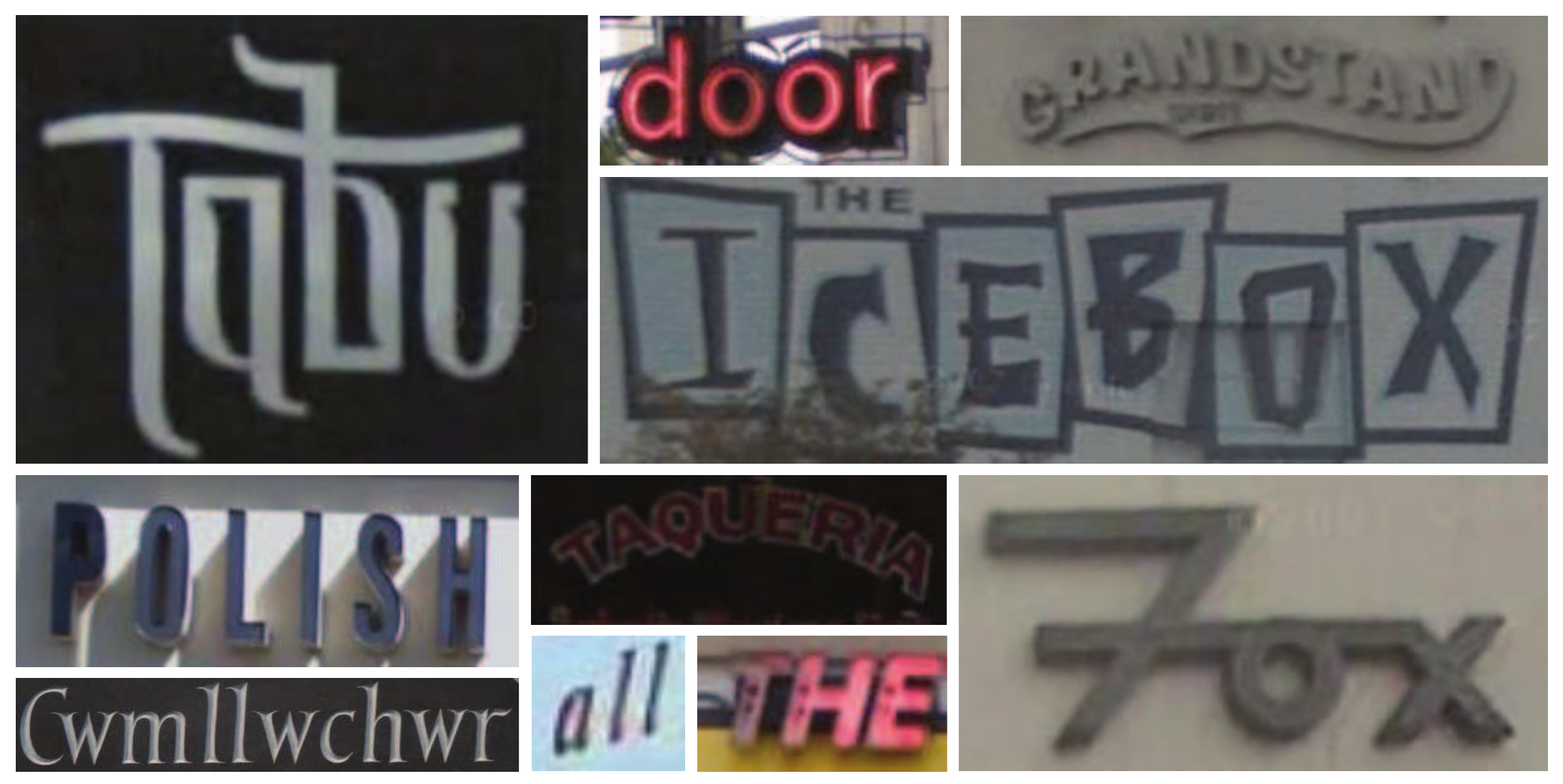}
    \end{center}
       \caption{Examples of difficult scene text images.}
    \label{fig1}
\end{figure}

Texts in the natural scene images convey rich and high-level semantic information which is important for image understanding. With the development of information technology, scene text recognition (STR) plays much more significant roles for image retrieval, intelligent transportation, robot navigation and so on. Consequently, STR has become a hot research topic in computer vision and artificial intelligence in recent years. Although the field of text recognition for scanned documents has observed tremendous progresses ~\cite{e1}~\cite{e2}, STR still remains a challenging problem, mainly due to the large variations in the aspects of background, resolution, text font and color, as shown in Fig. \ref{fig1}.

Many efforts have been devoted to attacking the difficulties of scene text recognition. Comprehensive surveys can be found in ~\cite{e20}~\cite{e35}. Traditional recognition schemes ~\cite{e8}~\cite{e3}~\cite{e11}~\cite{e7} usually adopt the bottom-up character detection or segmentation scheme. This approach has good interpretation since they can locate the position and label of each character. However, its performance is severely confined by the difficulty of character segmentation, and moreover, the method usually requires millions of manually labeled training samples for character classifier training (such as PhotoOCR ~\cite{e3}), which is both expensive and time-consuming.

Nowadays, deep neural network based frameworks have dominated the field of STR, by utilizing the top-down scheme which is independent of the segmentation performance and can be jointly optimized with the weakly labeled data. As the length of both input data and recognition result may vary drastically for STR, it is natural to make use of recurrent networks to capture context information and map them to an output sequence with variable length. In order to enhance the recognition performance, it is a common practice to combine CNN and RNN layers into hierarchical structures. RNN based method ~\cite{e14}~\cite{e19}%e12,
~\cite{e13}~\cite{e28}~\cite{f11} %, f11
is the dominant approach for dealing with the STR problem, and has obtained the state-of-the-art results on several challenging benchmarks. However, it mainly suffers two problems: (1) The training speed can be very slow as RNNs maintain a hidden state of the entire past that prevents parallel computation within a sequence; (2) The training process is tricky due to the gradient vanishing and exploding. Although the CNN based method can overcome the two defects of recurrent models, it is not widely used in the STR problem as CNNs often operate on the problems where the inputs and outputs are with fixed dimensions, and therefore they are incapable of producing a variable-length label sequence. A few methods \cite{e15}~\cite{e16} adopt the holistic methods with CNNs, which recognize words or text lines as a whole without character modeling.

\begin{figure}[tb]
    \begin{center}
    \includegraphics[width=0.8\linewidth]{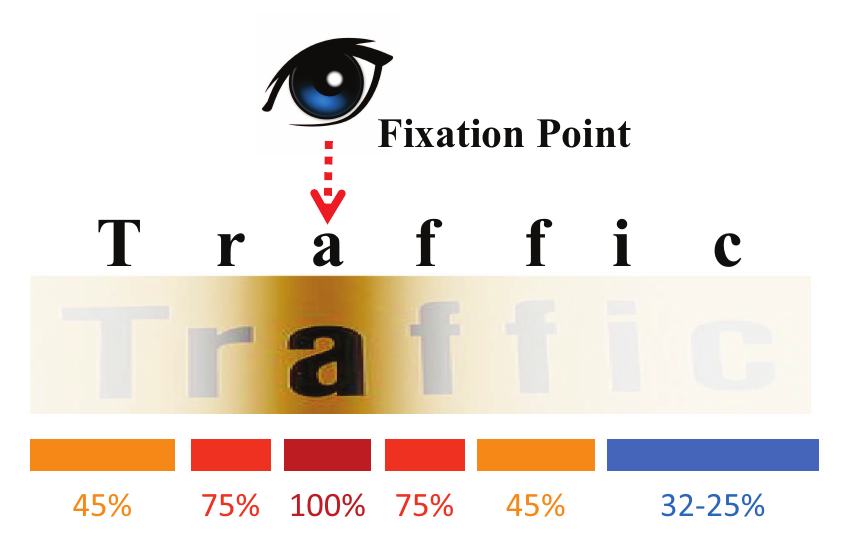}
    \end{center}
       \caption{Illustration of the acuity of foveal vision in reading. Around the fixation point only limited context information can been seen with 100\% acuity while reading the word "Traffic".}
    \label{fig2}
\end{figure}

On the other hand, there has been criticism that current state-of-the-art methods are usually applied in a black box manner and lack of interpretability for further improvement. Although the recurrent attention model can focus on part of the character, it is still unable to locate the boundary of each character. Modern cognitive psychology research points out that reading consists of a series of saccades and fixations ~\cite{e18}. Eyes do not move continuously along a line of text, but make short, rapid movements (saccades) intermingled with short stops (fixations). A diagram demonstrating the acuity of foveal vision in reading is shown in Fig. \ref{fig2}. Inspired by the above observations, and in order to realize this kind of biology phenomena into a computing model, we propose a novel method called SCAN (the abbreviation of Sliding Convolutional Attention Network) for scene text recognition with the purpose of interpretation and high performance.

The major contributions of this work are in three respects. First, we investigate the intrinsic characteristics of text recognition, and propose a novel scene text recognition method to imitate the mechanisms of saccades and fixations. Second, the model we propose for STR is entirely convolutional, which can be fully parallelized during training to better exploit the GPU hardware and optimization is easier since the number of non-linearities is fixed and independent of the input length. Third, we conduct extensive experiments on several benchmarks to demonstrate the superiority of the proposed method over state-of-the-art methods. Because of the good interpretability and flexibility of SCAN, we may easily utilize existing sophisticated network structures and post-processing techniques for further improvement. In addition, to the best of our knowledge, this is the first successful work to make use of convolutional sequence learning in the field of scene text recognition.

The work of ~\cite{e17} is close to ours in that sliding convolutional character model is used. However, this method still lacks of transparency, as it is a CTC based method, and its performance is not competitive compared with best systems. Different from the existing approaches, in this study, we combine the merits of both bottom-up and top-down methods, and propose an entirely convolutional model for faster training, better interpretability and higher performance.

The rest of this paper is organized as follows: Section \ref{proposed_model} gives a detailed introduction of the proposed method; Section \ref{experiment} presents experimental results, and Section \ref{conclusion} offers concluding remarks.

\section{Proposed Model}
\label{proposed_model}

\begin{figure}[ht]
    \begin{center}
    \includegraphics[width=1.0\linewidth]{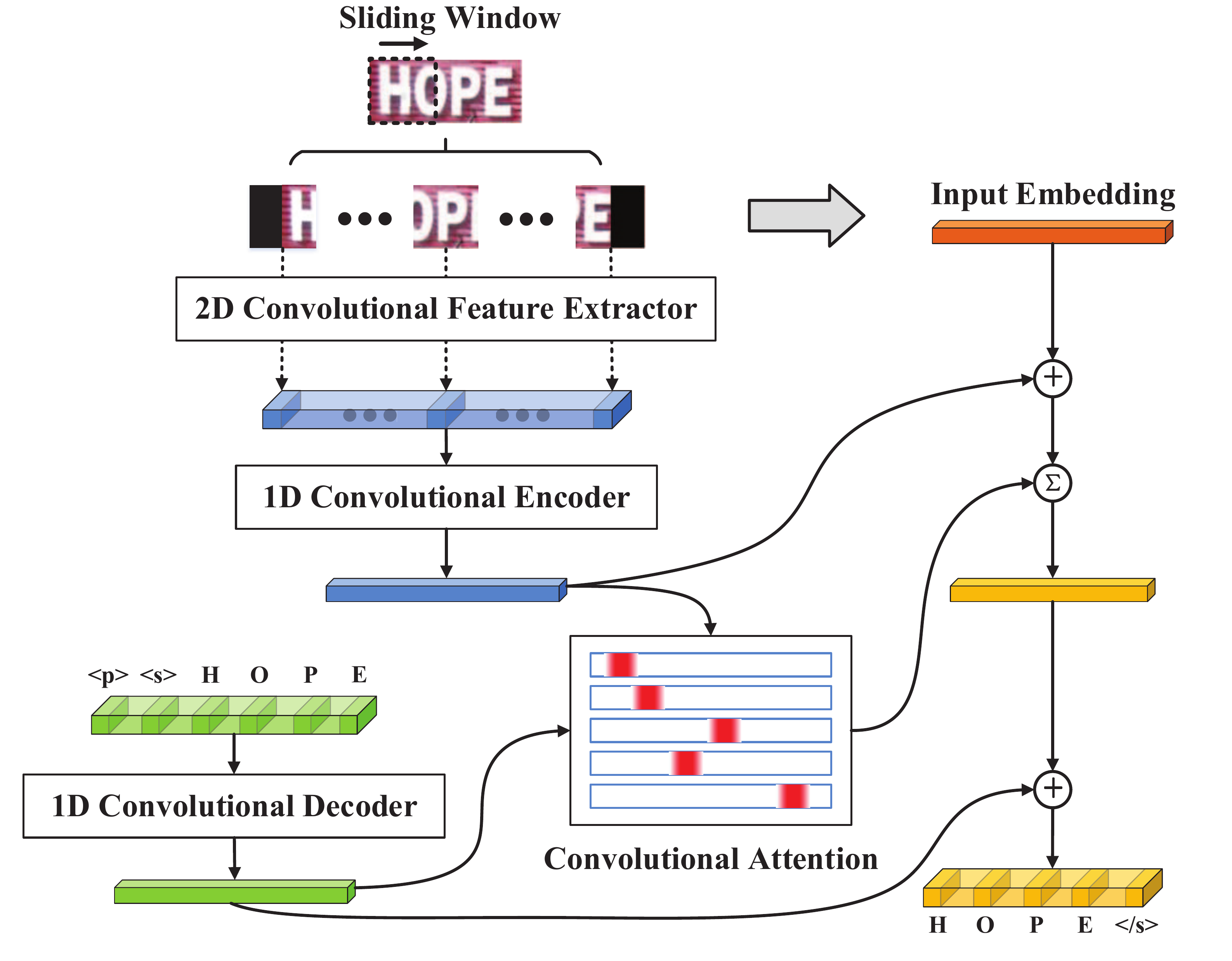}
    \end{center}
       \caption{The framework of SCAN. It consists of three parts: a sliding window layer, a convolutional feature extractor, and a convolutional sequence to sequence learning network.}
    \label{fig3}
\end{figure}

The SCAN recognition system is diagrammed in Fig. \ref{fig3}. It consists of three major parts, namely a sliding window layer, a convolutional feature extractor, and a convolutional sequence to sequence learning module including both the convolutional encoder and the convolutional decoder.

The sliding window layer splits the textline into overlapped windows. On the top of sliding window, a convolutional feature extractor is built to extract the discriminative features. Finally, a convolutional sequence to sequence learning module is adopted to transform feature sequence into the result sequence. The whole system can be fully parallelized and jointly optimized independent of the input length, as all these three layers can be processed simultaneously while training.

\subsection{Multi-Scale Sliding Window}

When humans read a text line, their eyes do not move continuously along a line of text, but make short rapid movements intermingled with short stops. During the time that the eye is stopped, new information is brought into the processing, but during the movements, the vision is suppressed so that no new information is acquired ~\cite{e18}. Inspired by this, we use exhaustive scan windows with suitable step to imitate the saccade, the centre of the scan window could be a potential fixation point, while the perceptual span can be referred to the window size. Therefore, this mechanism can be simplified to the sliding window.

\begin{figure}[tb]
    \subfigure[Characters with various width]{
    \label{fig5a}
    \includegraphics[width=0.21\textwidth]{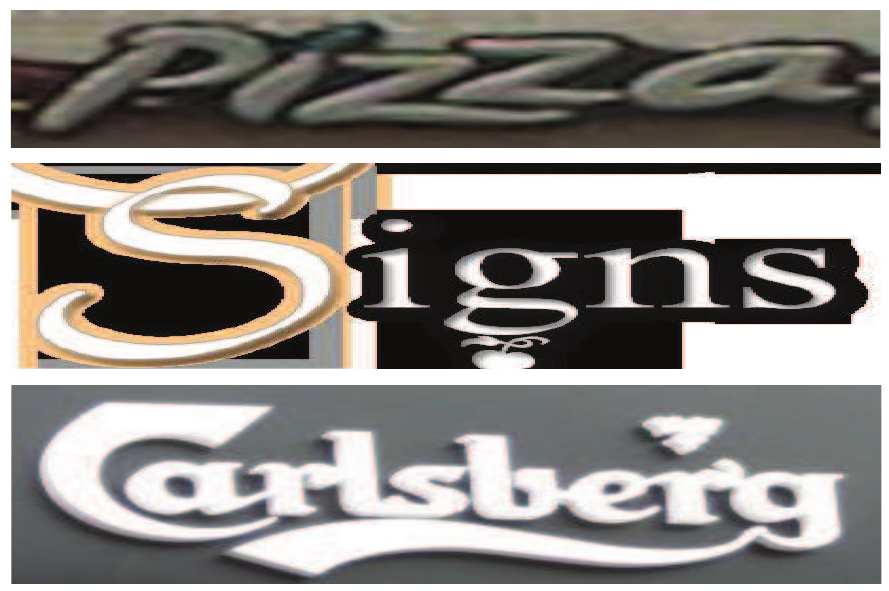}
    }
    \subfigure[Multi-scale sliding window]{
    \includegraphics[width=0.24\textwidth]{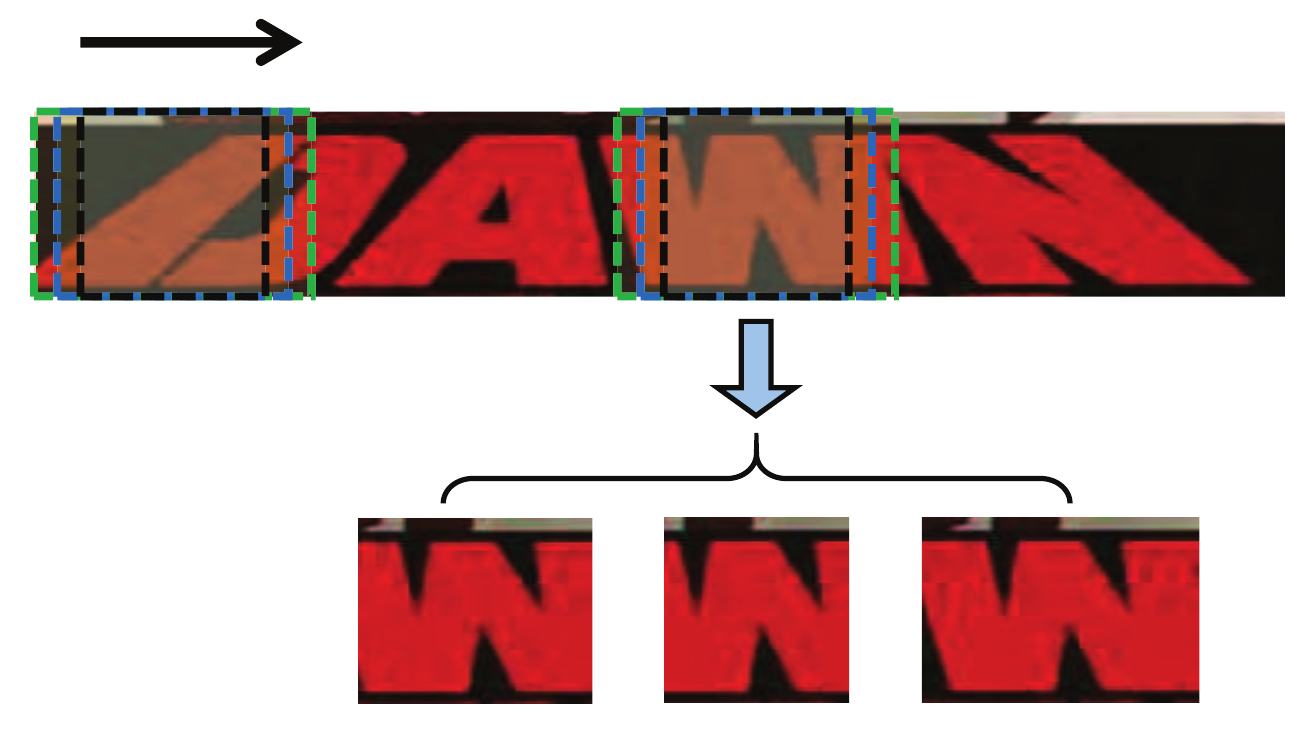}
    \label{fig5b}
    }
    \caption{Illustration of multi-scale sliding window mechanism. The model can take different glimpses to adjust different character fonts and capture more accurate context information.}
\end{figure}

If we assume the model only skips one character in each saccade as a unskilled people, we could fix the sliding window to the size where a character can be covered completely. Although after height normalization, characters usually have approximately similar width in the text image, the aspect ratio of the character may still vary between different fonts (Fig. \ref{fig5a}). Thus, we propose to use the sliding window with multi scales, as shown in Fig \ref{fig5b}. The model can then take different glimpses to adjust different character fonts and capture more accurate context information. Those different glimpses will be proportionally resized to the same size, and then concatenated together as multi-channel inputs which will be fed to the convolutional feature extractor.

\subsection{Convolutional Feature Extraction}
\label{conv_extraction}

\begin{figure}[tb]
    \begin{center}
    \includegraphics[width=0.98\linewidth]{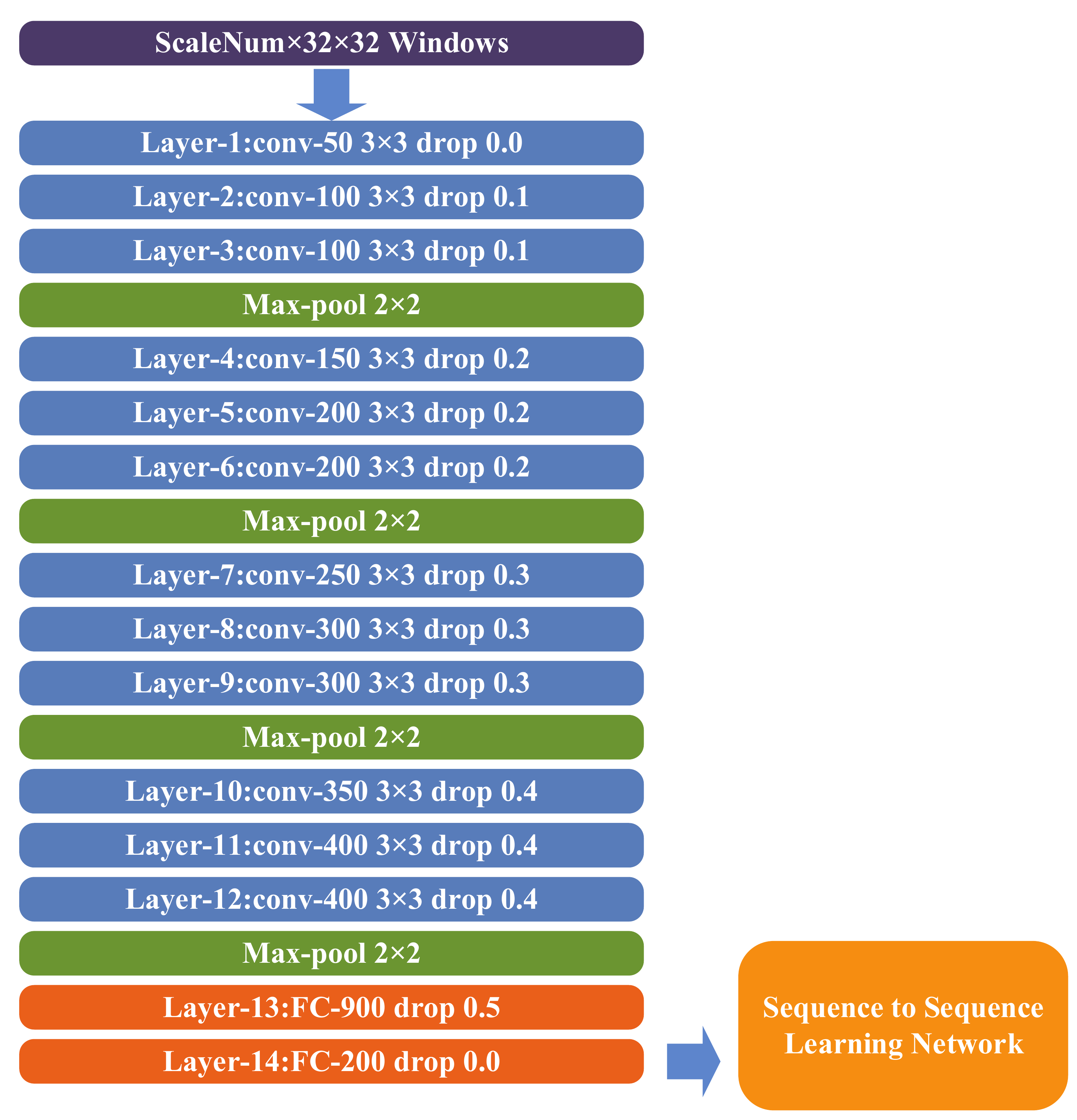}
    \end{center}
    \caption{The architecture for convolutional feature extraction.}
    \label{fig5x}
\end{figure}

Although we may use any feature extraction method (such as HOG ~\cite{e22}) to embed the resized windows in distributional space, we adopt the CNN based approach for its superior performance in the field of pattern recognition and its ability of easy integrating into other modules for end-to-end training. We build a 14-layer network for convolutional feature extraction \footnote{We may easily adopt existing sophisticated network structures for further improvement.}, as shown in Fig. \ref{fig5x}. The structure is inherited from ~\cite{e2}, while we remove the Softmax layer of the original network. In this way, the resized window can be finally reduced to a 200-dimensional high-level representation.

%\subsection{Sequence to Sequence Learning Network}
\subsection{Convolutional Sequence Learning}
After convolutional feature extraction, we obtain a deep feature sequence representing the information captured from a series of windows, which can be denoted as $\bm{s} = (s_1, ..., s_m)$. The process of fixations can be formulated by a convolutional sequence to sequence learning module to generate the recognition result $\bm{y} = (y_1, ..., y_n)$.

The recurrent neural network based encoder-decoder architectures ~\cite{e29} have dominated the field of sequence to sequence learning. To make full use of the context information, it is better to use architectures with attention ~\cite{e30}, which compute the conditional input $c_i$ as a weighted sum of encoder state representations $(z_1, ..., z_m)$ at each time step. The weights of the sum are referred to as attention scores and allow the network to focus on different parts of the input sequence as it generates the output sequences, while in this work, we interpret it as the mechanism of fixation. As for the STR problem, all the previous attention based frameworks employ the recurrent models to model long-term dependencies.

Instead of relying on RNNs to compute intermediate encoder states $\bm{z}$ and decoder states $\bm{h}$, we turn to use a fully convolutional architecture for sequence to sequence modeling shown in Fig. \ref{fig3}. Gehring et al. ~\cite{e32} propose the convolutional sequence to sequence learning architecture for the first time, which is successfully applied to the machine translation problem with better accuracy and faster training speed.

In order to equip our model with a sense of order, the input feature representation $\bm{s} = (s_i, ..., s_m)$ is combined with the absolute position embedding of input elements $\bm{p} = (p_1, ..., p_m)$ to obtain the input element representations $\bm{e} = (s_1 + p_1, ..., s_m + p_m)$. We proceed similarly for output elements, denoted as $\bm{g} = (g_1, ..., g_n)$.

For convolutional sequence learning, both encoder and decoder networks compute intermediate states based on a fixed number of input elements using a simple layer structure. We denote the output of the $l$-th layer as $z^l = (z^l_1, ..., z^l_m )$ for the encoder network, and $h^l = (h^l_1, ..., h^l_n)$ for the decoder network. A one dimensional convolution exists in each layer followed by a non-linearity. For a network with a single layer and kernel width $k$, each resulting state $h^l_i$ contains information over $k$ input elements. Stacking several blocks on top of each other increases the number of input elements represented in a state. The final distribution over the possible next target elements $y_{i+1}$ can be calculated by transforming the top decoder output $h^L_i$ via a linear layer with weights $W_o$ and bias $b_o$:
\begin{equation}\label{fun_distribution}
p(y_{i + 1}|y_1, ..., y_i, \bm{s})=\text{Softmax}(W_oh^L_i + b_o).
\end{equation}

The convolutional sequence learning network adopts a separate attention mechanism for each decoder layer. To compute the attention vector, the current decoder state $h^l_i$ is combined with an embedding of the previous target element $g_i$:
\begin{equation}\label{fun_attention_1}
d^l_i = W^l_dh^l_i + b^l_d + g_i.
\end{equation}
For decoder layer $l$, the attention $a^l_{ij}$ of state $i$ is a dot-product between the decoder state summary $d^l_i$ and each output $z^u_j$ of the last encoder block $u$:
\begin{equation}\label{fun_attention_2}
a^l_{ij}=\frac{\text{exp}(d^l_i \cdot z^u_j)}{\sum^m_{t = 1}{\text{exp}(d^l_i \cdot z^u_t)}}.
\end{equation}
The input to the current decoder layer, denoted as $c^l_i$, is computed as a weighted sum of the encoder outputs as well as the input element embeddings $e_j$:
\begin{equation}\label{fun_attention_3}
c^l_i=\sum^m_{j = 1}{a^l_{ij}(z^u_j + e_j)},
\end{equation}
where encoder output $z^u_j$ represents potentially large input contexts, and input embedding $e_j$ provides point information about a specific input element that is useful when making a prediction. Then we added $c^l_i$ to the output of the corresponding decoder layer $h^l_i$ once it has been computed.

For multi-step attention mechanism, the attention of the first layer determines a useful source context which is then fed to the second layer that takes this information into account when computing attention. This makes it easier for the model to take into account which previous inputs have been attended to already. While for recurrent nets, this information is in the recurrent state and needs to survive several non-linearities. The convolutional architecture also allows to batch the attention computation across all elements of a sequence compared to RNNs.

\subsection{Model Training}

Denote the training dataset by $D = \{X_i, Y_i\}$, where $X_i$ is a training image of textline, $Y_i$ is a ground truth label sequence. To train the model, we minimize the negative log-likelihood over $D$:
\begin{equation}\label{fun_train}
\mathcal{L}=-\sum_{S_i, Y_i \in D}{\log{p(Y_i|S_i)}},
\end{equation}
where $S_i$ is the window sequence produced by sliding on the image $X_i$. This objective function calculates a cost value directly from an image and its ground truth label sequence. Therefore, the network can be end-to-end trained on pairs of images and sequences by the standard back-propagation algorithm, eliminating the procedure of manually labeling all individual characters in training images. This is a weakly supervised learning manner.

\subsection{Decoding}

The convolutional decoder network is to generate the output sequence of characters from the implicitly learned character-level probability statistics. In the process of unconstrained text recognition (lexicon-free), we use a beam of width $K$ to select the most probable character sequence. As soon as the ``$\langle /s \rangle$'' symbol is appended to a hypothesis, it is removed from the beam and is added to the set of complete hypotheses. In order to overcome the bias to short strings, we divide the log-likelihoods of the final hypothesis in beam search by their length $|\bm{y}|$.

While in lexicon-driven text recognition, firstly, we select the top $K$ hypothesis using beam search, and then calculate the Levenshtein distance between each hypotheses and each item of the lexicon. We choose the one with the shortest distance as the recognition result. We set $K$ to be 5 in this work. It should be mentioned that we also develop a simple version of decoding algorithm with prefix tree. We abandon it because we do not find any improvement compared with the forementioned straightforward method.

\section{Experiments}
\label{experiment}

In this section we evaluate the proposed SCAN model on four standard scene text recognition benchmarks. Firstly, we give the experiment settings and implementation details. Then, we analyze the design choices in SCAN. For comprehensive performance comparison, SCAN is compared with existing typical methods. For all benchmarks, the performance is measured by word-level accuracy.

\subsection{Datasets}

We use the synthetic dataset released by ~\cite{f3} as training data for all the following experiments. The training set consists of 8 millions images and their corresponding ground truth on text line level. Although we only used the synthetic data to train our model, even without any fine tuning on specific training sets, it works well on real image datasets. We evaluated our scene text recognition system on four popular benchmarks, namely IIIT 5k-word (IIIT5k) ~\cite{e4}, Street View Text (SVT) ~\cite{e8}, ICDAR 2003 (IC03) ~\cite{f1}, and ICDAR 2013 (IC13) ~\cite{f2}.

\subsection{Implementation Details}

During training, all images are normalized to $32\times256$ for parallel computation on GPU. If the proportional width is less than 256, we pad the scaled image to width 256, otherwise, we continue to scale the image to $32\times256$ (this rarely happens because most words are not so long). We do the same normalization on testing images to evaluate the effectiveness of SCAN.

We used 256 hidden units for both convolutional encoders and decoders. We trained networks using Adam ~\cite{f4} with the learning rate of 0.0005 and renormalized the gradients if their norm exceeded 0.1. We used mini-batches of 40 images and selected 1\% of the training samples for each epoch. Meanwhile, the dropout was set to be 0.5 on both encoder and decoder.

We implemented the model on the platform of Torch 7 ~\cite{a167}, and carefully designed our system so that the models can be trained on multiple GPUs simultaneously. Experiments were performed on a workstation with the Intel(R) Xeon(R) E5-2680 CPU, 256GB RAM and four NVIDIA GeForce GTX TITAN X GPUs. It took about only 10 to 15 minutes per epoch, and the training can usually be finished after about 500 epochs\footnote{The model can usually converge to relatively good results in only about 150 epochs.}. In fact, we also implemented a recurrent sequence learning version of SCAN in private. We found that SCAN in this study is at least 9 times faster than the recurrent model in training speed, which can be attributed to the fully parallelization of the proposed model. SCAN takes only 0.3s to recognize an image on average. We could further improve the speed with better implementation.

\subsection{Experimental Results}
To our knowledge, SCAN is the first work to adopt convolutional sequence learning network in the field of STR. We find the recommended hyperparameters in \cite{e32} do not work well, thus, we investigate the effects of some major parameters of SCAN.

\subsubsection{Kernel size and Depth}

Table \ref{T1} and \ref{T2} shows recognition accuracies when we change the number of layers in the encoder or decoder. Here the kernel widths are fixed to be 5 and 7 for layers in the encoder and decoder, respectively. Deeper architectures are particularly beneficial for the encoder but less so for the decoder. Decoder setups with two layers already perform well, whereas for the encoder accuracy keeps increasing steadily to three layers. In fact, we have tried to further increase the number of encoder layers, but find there exists a severe overfitting problem.

\begin{table}[ht]
\begin{center}
\begin{tabular}{ccccc}
\hline
Encoder Layer & IIIT5k & SVT & IC03 & IC13\\
\hline
\multicolumn{1}{c|}{1} & 83.6 & 81.0 & 90.1 & 89.6\\
\multicolumn{1}{c|}{2} & 84.0 & 81.1 & 90.3 & \textbf{90.3}\\
\multicolumn{1}{c|}{3} & \textbf{84.9} & \textbf{82.4} & \textbf{91.4} & 90.1\\
\hline
\end{tabular}
\end{center}
\caption{Effects of encoder layers with 2 decoder layers in terms of accuracy (\%).}
\label{T1}
\end{table}

\begin{table}[ht]
\begin{center}
\begin{tabular}{ccccc}
\hline
Decoder Layer & IIIT5k & SVT & IC03 & IC13\\
\hline
\multicolumn{1}{c|}{1} & 82.9 & 81.5 & 90.3 & 89.0\\
\multicolumn{1}{c|}{2} & \textbf{84.9} & 82.4 & \textbf{91.4} & \textbf{90.1}\\
\multicolumn{1}{c|}{3} & 84.2 & \textbf{83.5} & 90.5 & 89.4\\
\hline
\end{tabular}
\end{center}
\caption{Effects of decoder layers with 3 encoder layers in terms of accuracy (\%). }
\label{T2}
\end{table}

\begin{table}[ht]
\begin{center}
\begin{tabular}{ccccc}
\hline
Encoder-Decoder & IIIT5k & SVT & IC03 & IC13\\
\hline
\multicolumn{1}{c|}{5-5} & 84.6 & \textbf{82.5} & 90.5 & 89.4\\
\multicolumn{1}{c|}{5-7} & \textbf{84.9} & 82.4 & \textbf{91.4} & \textbf{90.1}\\
\multicolumn{1}{c|}{7-7} & 84.1 & \textbf{82.5} & 89.9 & 88.8\\
\hline
\end{tabular}
\end{center}
\caption{Effects of kernel width in terms of accuracy (\%).}
\label{T3}
\end{table}

Aside from increasing the depth of the networks, we also change the kernel width. The recognition accuracies with different kernel widths are listed in Table \ref{T3}. The kernel size of encoders can be referred to the perceptual span across different windows, and the width of 5 performs the best, which is a little larger than the width of a character. The decoder networks can be seen as a language model, it is better to choose large kernel sizes.

\subsubsection{Multi-Scale Sliding Window}

We investigate two types of model: single-scale model and multi-scale model. The single-scale model has only one input feature map with the window size of $32\times32$, $32\times40$ or $32\times48$. While the multi-scale model has three input feature maps, which are firstly extracted with all the three forementioned window sizes and then concatenated together. All the windows are resized to $32\times32$ before fed to the convolutional feature extractor. The sliding window is shifted with the step of 4.

\begin{table}[htb]
\begin{center}
\begin{tabular}{ccccc}
\hline
Scale-model & IIIT5k & SVT & IC03 & IC13\\
\hline
\multicolumn{1}{c|}{n=1 ($32\times32$)} & 83.7 & 82.1 & 89.9 & 89.3\\
\multicolumn{1}{c|}{n=1 ($32\times40$)} & \textbf{84.9} & 82.4 & 91.4 & 90.1\\
\multicolumn{1}{c|}{n=1 ($32\times48$)} & 84.5 & 82.1 & \textbf{92.1} & 90.0\\
\multicolumn{1}{c|}{n=3 }               & 84.2 & \textbf{85.0} & \textbf{92.1} & \textbf{90.4}\\
\hline
\end{tabular}
\end{center}
\caption{Effects of scale models in terms of accuracy (\%).}
\label{T4}
\end{table}

The recognition accuracies of models with different scales are shown in Table \ref{T4}. As for the single-scale model, it can be seen that the $32 \times 40$ window achieve the best performance. The window with large span can deal with more information, which is beneficial for SCAN. However, if the window size is too large (such as $32 \times 48$), it may bring many disturbances for feature extraction. The multi-scale model is significantly better than the single one, as it can then take different glimpses to adjust different character fonts and capture more accurate context information.

\subsection{Comparative Evaluation}

\begin{table*}[htb]
%\footnotesize
\scriptsize
\centering
\caption{Recognition accuracies (\%) on four standard scene text datasets. In the second row, 50, 1k and Full denote the lexicon used, and None denotes recognition without language constraints.(* %~\cite{e16}
 is not lexicon-free in the strict sense, as its outputs are constrained to a 90k dictionary.)}
\label{TRST}
\begin{tabular}{cccccccccccccc}%{0.9\textwidth}
\hline
  \multirow{2}{*}{Method}                                 & \multicolumn{3}{c}{IIIT5k} &  & \multicolumn{2}{c}{SVT} &  & \multicolumn{3}{c}{IC03} &  & IC13 \\
                                                            \cline{2-4} \cline{6-7} \cline{9-11} \cline{13-13}
                                                          & 50    & 1k    & None    &  & 50         & None       &  & 50     & Full   & None   &  & None \\ \hline
  \multicolumn{1}{c|}{ABBYY \cite{e8}}                    & 24.3  & -     & -       &  & 35.0       & -          &  & 56.0   & 55.0   & -      &  & -    \\
  \multicolumn{1}{c|}{Wang et al. \cite{e8}}              & -     & -     & -       &  & 57.0       & -          &  & 76.0   & 62.0   & -      &  & -    \\
  \multicolumn{1}{c|}{Mishra et al. \cite{e4}}            & 64.1 & 57.5 & -    &  & 73.2 & -     &  & 81.8 & 67.8 & -     &  & -         \\
  \multicolumn{1}{c|}{Novikova et al. \cite{e6}}          & -    & -    & -    &  & 72.9 & -     &  & 82.8 & -    & -     &  & -         \\
  \multicolumn{1}{c|}{Wang et al. \cite{e9}}              & -    & -    & -    &  & 70.0 & -     &  & 90.0 & 84.0 & -     &  & -         \\
  \multicolumn{1}{c|}{Bissaco et al. \cite{e3}}           & -    & -    & -    &  & 90.4 & 78.0  &  & -    & -    & -     &  & 87.6      \\
  \multicolumn{1}{c|}{Goel et al. \cite{f5}}              & -    & -    & -    &  & 77.3 & -     &  & 89.7 & -    & -     &  & -         \\
  \multicolumn{1}{c|}{Alsharif \& Pineau \cite{f6}}       & -    & -    & -    &  & 74.3 & -     &  & 93.1 & 88.6 & -     &  & -         \\
  \multicolumn{1}{c|}{Almazan et al. \cite{f7}}           & 91.2 & 82.1 & -    &  & 89.2 & -     &  & -    & -    & -     &  & -         \\
  \multicolumn{1}{c|}{Yao et al. \cite{e11}}              & 80.2 & 69.3 & -    &  & 75.9 & -     &  & 88.5 & 80.3 & -     &  & -         \\
  \multicolumn{1}{c|}{R.-Serrano et al. \cite{f8}}        & 76.1 & 57.4 & -    &  & 70.0 & -     &  & -    & -    & -     &  & -         \\
  \multicolumn{1}{c|}{Su \& Lu et al. \cite{e12}}         & -    & -    & -    &  & 83.0 & -     &  & 92.0 & 82.0 & -     &  & -         \\
  \multicolumn{1}{c|}{Gordo \cite{f9}}                    & 93.3 & 86.6 & -    &  & 91.8 & -     &  & -    & -    & -     &  & -         \\
  \multicolumn{1}{c|}{Jaderberg et al. \cite{e16}}        & 97.1 & 92.7 & -    &  & 95.4 & 80.7* &  & \textbf{98.7} & \textbf{98.6} & \textbf{93.1*} &  & \textbf{90.8*}   \\
  \multicolumn{1}{c|}{Jaderberg et al. \cite{e15}}        & 95.5 & 89.6 & -    &  & 93.2 & 71.7  &  & 97.8 & 97.0 & 89.6  &  & 81.8    \\
  \multicolumn{1}{c|}{Shi et al. \cite{e13}}              & 97.8 (5) & 95.0 (5) & 81.2 (6) &  & \textbf{97.5 (1)} & 82.7 (2) &  & \textbf{98.7 (1)} & 98.0 (1)& 91.9 (2) &  & 89.6 (4)  \\
  \multicolumn{1}{c|}{Shi et al. \cite{e14}}             & 96.2 (8) & 93.8 (8)& 81.9 (4)&  & 95.5 (5)& 81.9 (5) &  & 98.3 (3)& 96.2 (7)& 90.1 (5) &  & 88.6 (6)      \\
  \multicolumn{1}{c|}{Lee et al. \cite{e28}}              & 96.8 (7) & 94.4 (7)& 78.4 (8)&  & 96.3 (2)& 80.7 (6) &  & 97.9 (5)& 97.0 (3)& 88.7 (7) &  & 90.0 (3)     \\
  \multicolumn{1}{c|}{Yin et al. \cite{e17}}              & 98.9 (2) & 96.7 (4)& 81.6 (5)&  & 95.1 (7)& 76.5 (8) &  & 97.7 (7)& 96.4 (6)& 84.5 (8) &  & 85.2 (8)  \\
  \multicolumn{1}{c|}{Cheng et al. \cite{e19} (baseline)} & 98.9 (2)& 96.8 (3)& 83.7 (3)&  & 95.7 (3)& 82.2 (4) &  & 98.5 (2)& 96.7 (5)& 91.5 (3) &  & 89.4 (5)      \\ \hline
  \multicolumn{1}{c|}{Ours (n=1)}                          & 98.8 (4)& 97.1 (2)& \textbf{84.9 (1)} &  & 95.5 (5)& 82.4 (3) &  & 97.8 (6)& 97.0 (3)& 91.4 (4) &  & 90.1 (2)  \\
  \multicolumn{1}{c|}{Ours (n=3)}                          & \textbf{99.1 (1)} & \textbf{97.2 (1)} & 84.2 (2)&  & 95.7 (3)& \textbf{85.0 (1)}  &  & 98.3 (3)& 97.2 (2)& \textbf{92.1 (1)}&& \textbf{90.4 (1)}   \\
  \multicolumn{1}{c|}{Ours (n=1, Residual)}                 & 97.8 (5) & 94.5 (6)& 79.8 (7)&  & 94.7 (8)& 77.4 (7) &  & 97.4 (8)& 95.7 (8)& 89.2 (6) &  & 87.7 (7)  \\ \hline
\end{tabular}
\end{table*}

We choose the combination of the parameters that achieve the best performance in the above discussion, and evaluate our model on four public datasets of scene text word recognition. The results are listed in Table \ref{TRST} with comparison to state-of-the-art methods. It should be mentioned that although Cheng et al. ~\cite{e19} report the best performance with ResNet-based network and focusing attention, they train the model with much more training data (an extra of 4-million pixel-wise labeled word images). Therefore, we only list the baseline performance in their work for fair comparison, which is a ResNet-based structure.

\begin{figure}[htb]
    \begin{center}
    \subfigure[Correct recognition samples]{
    \label{fig7a}
    \includegraphics[width=0.4\textwidth]{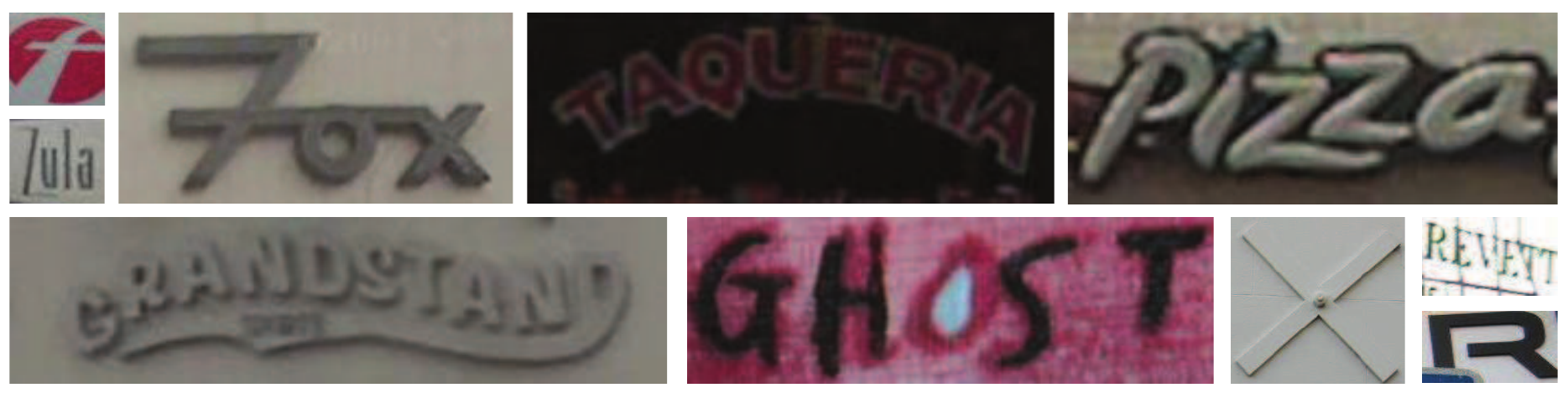}
    }

    \subfigure[Incorrect recognition samples]{
    \includegraphics[width=0.4\textwidth]{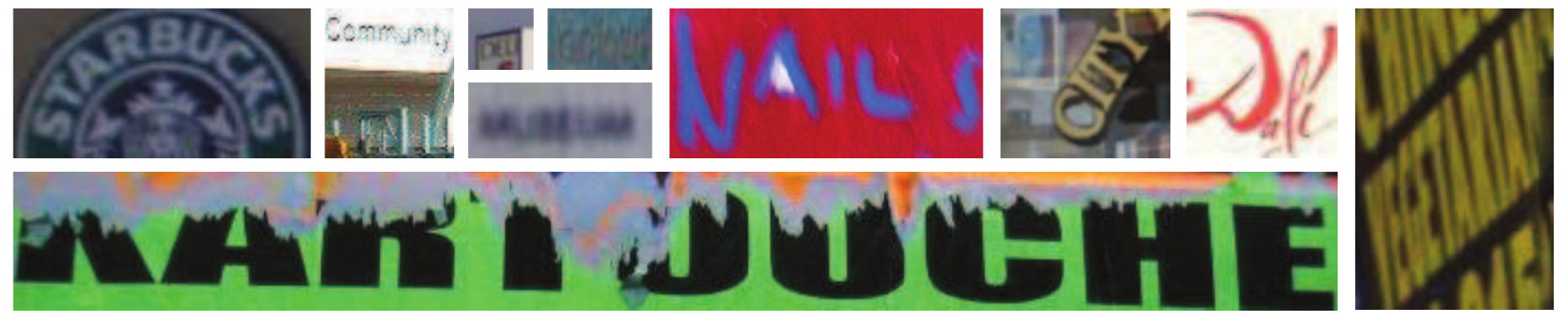}
    \label{fig7b}
   }
    \end{center}
    \caption{Recognition examples of SCAN.}
    \label{fig7}
\end{figure}

In the lexicon-free case (None), our model outperforms all the other methods in comparison. On IIIT5k, SCAN outperforms prior art ~\cite{e19} by nearly 1\%, while on SVT set, it outperforms prior art CRNN ~\cite{e13} by significantly 2.3\%, indicating a clear improvement in performance. ~\cite{e19}~\cite{e13} are both recurrent model based works, while with convolutional sequence learning, SCAN achieves state-of-the-art among the methods using the same data. We observe that IIIT5k contains a lot of irregular text, especially curved text. Although SCAN takes a simple sliding window scheme, because of the powerful feature extractor and sequence learning network, it has an advantage in dealing with irregular text over the complicated RARE model ~\cite{e14}. In the constrained lexicon cases, our method consistently outperforms most state-of-the-arts approaches, and in average beats best text readers proposed in ~\cite{e13}~\cite{e14}~\cite{e19}. The average rank including works ~\cite{e13}~\cite{e14}~\cite{e28}~\cite{e17}~\cite{e19} and ours shows the superiority of SCAN over all the prior arts. Some recognition examples of SCAN are shown in Fig. \ref{fig7}.

\subsection{Visualization}

\begin{figure}[htb]
    \begin{center}
    \includegraphics[width=1.0\linewidth]{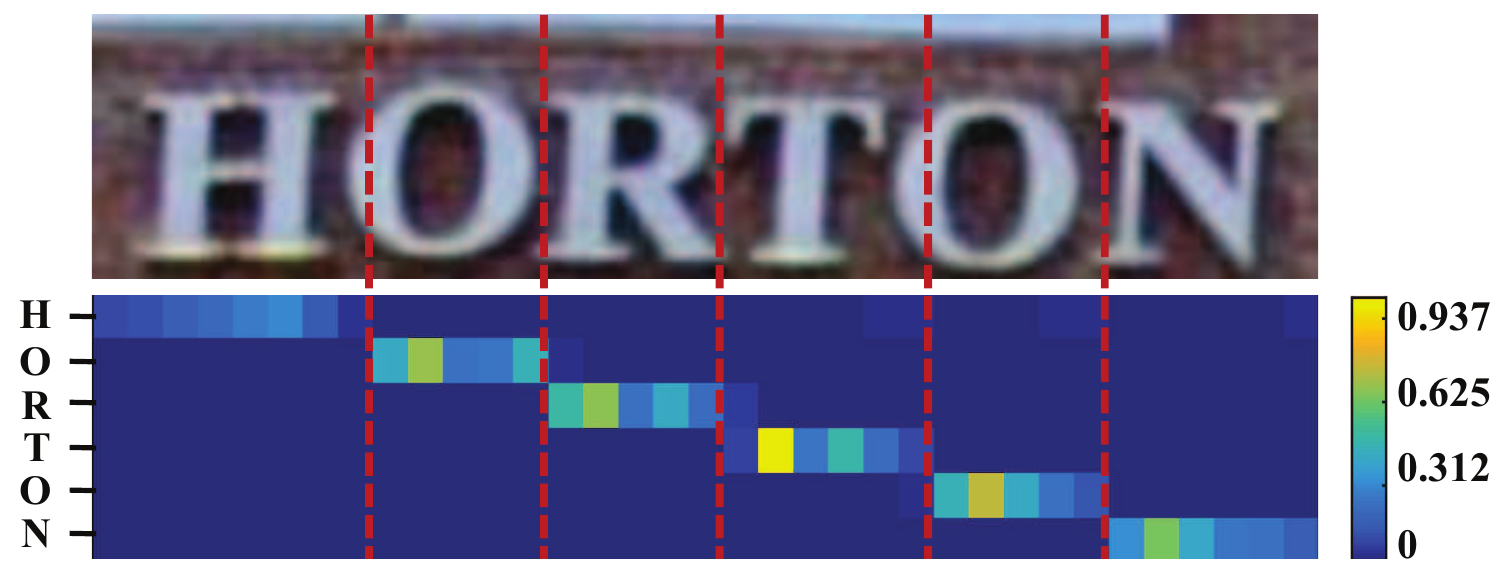}
    \end{center}
       \caption{An example of the attention weights while reading a text image containing the word "HORTON". Each point in the attention heatmap corresponds to the center of a sliding window.}
    \label{fig8}
\end{figure}

An example of the attention weights for the SCAN model is shown Fig. \ref{fig8}. There exist clear gaps between different characters on the attention heatmap, which reveals that the model can accurately attend on the most relevant sliding windows. Although those recurrent attention model can focus on part of the character ~\cite{e14}~\cite{e19}, they are unable to locate the boundary of each character. More visualization examples can be found in the supplemental materials.

\section{Conclusion and Future Work}
\label{conclusion}

In this paper, we take advantage of the intrinsic characteristics of text recognition, and inspired by human cognition mechanisms in reading texts, we propose a novel method of SCAN for scene text recognition with good interpretation and high performance. Different from the existing approaches, SCAN is an entirely convolutional model based on sliding window and sequence learning, and it acts very similar to the process of human reading. The experimental results on several challenging benchmarks demonstrate the superiority of SCAN over state-of-the-art methods in terms of both the model interpretability and performance. Future work will concentrate on further improvement of the model performance on more challenging datasets.

\bibliographystyle{named}
\bibliography{arXiv}

\begin{thebibliography}{}

\bibitem[\protect\citeauthoryear{Almaz{\'a}n \bgroup \em et al.\egroup
  }{2014}]{f7}
Jon Almaz{\'a}n, Albert Gordo, Alicia Forn{\'e}s, and Ernest Valveny.
\newblock Word spotting and recognition with embedded attributes.
\newblock {\em IEEE Trans. Pattern Analysis and Machine Intelligence},
  36(12):2552--2566, 2014.

\bibitem[\protect\citeauthoryear{Alsharif and Pineau}{2013}]{f6}
Ouais Alsharif and Joelle Pineau.
\newblock End-to-end text recognition with hybrid hmm maxout models.
\newblock {\em arXiv preprint arXiv:1310.1811}, 2013.

\bibitem[\protect\citeauthoryear{Bahdanau \bgroup \em et al.\egroup
  }{2015}]{e30}
Dzmitry Bahdanau, Kyunghyun Cho, and Yoshua Bengio.
\newblock Neural machine translation by jointly learning to align and
  translate.
\newblock In {\em Proc. ICLR}, 2015.

\bibitem[\protect\citeauthoryear{Bissacco \bgroup \em et al.\egroup
  }{2013}]{e3}
Alessandro Bissacco, Mark Cummins, Yuval Netzer, and Hartmut Neven.
\newblock Photoocr: Reading text in uncontrolled conditions.
\newblock In {\em Proc. ICCV}, pages 785--792, 2013.

\bibitem[\protect\citeauthoryear{Cheng \bgroup \em et al.\egroup }{2017}]{e19}
Zhanzhan Cheng, Fan Bai, Yunlu Xu, Gang Zheng, Shiliang Pu, and Shuigeng Zhou.
\newblock Focusing attention: Towards accurate text recognition in natural
  images.
\newblock In {\em Proc. ICCV}, pages 5076--5084, 2017.

\bibitem[\protect\citeauthoryear{Collobert \bgroup \em et al.\egroup
  }{2011}]{a167}
Ronan Collobert, Koray Kavukcuoglu, and Cl{\'e}ment Farabet.
\newblock Torch7: A {MATLAB}-like environment for machine learning.
\newblock In {\em Proc. NIPS Workshop of BigLearn}, number EPFL-CONF-192376,
  2011.

\bibitem[\protect\citeauthoryear{Dalal and Triggs}{2005}]{e22}
Navneet Dalal and Bill Triggs.
\newblock Histograms of oriented gradients for human detection.
\newblock In {\em Proc. CVPR}, volume~1, pages 886--893, 2005.

\bibitem[\protect\citeauthoryear{Gehring \bgroup \em et al.\egroup
  }{2017}]{e32}
Jonas Gehring, Michael Auli, David Grangier, Denis Yarats, and Yann~N Dauphin.
\newblock Convolutional sequence to sequence learning.
\newblock In {\em Proc. ICML}, 2017.

\bibitem[\protect\citeauthoryear{Goel \bgroup \em et al.\egroup }{2013}]{f5}
Vibhor Goel, Anand Mishra, Karteek Alahari, and CV~Jawahar.
\newblock Whole is greater than sum of parts: Recognizing scene text words.
\newblock In {\em Proc. 12th Int. Conf. on Document Analysis and Recognition},
  pages 398--402, 2013.

\bibitem[\protect\citeauthoryear{Gordo}{2015}]{f9}
Albert Gordo.
\newblock Supervised mid-level features for word image representation.
\newblock In {\em Proc. CVPR}, pages 2956--2964, 2015.

\bibitem[\protect\citeauthoryear{Graves \bgroup \em et al.\egroup }{2009}]{e1}
Alex Graves, Marcus Liwicki, Santiago Fern{\'a}ndez, Roman Bertolami, Horst
  Bunke, and J{\"u}rgen Schmidhuber.
\newblock A novel connectionist system for unconstrained handwriting
  recognition.
\newblock {\em IEEE Trans. Pattern Analysis and Machine Intelligence},
  31(5):855--868, 2009.

\bibitem[\protect\citeauthoryear{Jaderberg \bgroup \em et al.\egroup
  }{2014}]{f3}
Max Jaderberg, Karen Simonyan, Andrea Vedaldi, and Andrew Zisserman.
\newblock Synthetic data and artificial neural networks for natural scene text
  recognition.
\newblock {\em arXiv preprint arXiv:1406.2227}, 2014.

\bibitem[\protect\citeauthoryear{Jaderberg \bgroup \em et al.\egroup
  }{2015}]{e16}
Max Jaderberg, Karen Simonyan, Andrea Vedaldi, and Andrew Zisserman.
\newblock Deep structured output learning for unconstrained text recognition.
\newblock In {\em Proc. ICLR}, 2015.

\bibitem[\protect\citeauthoryear{Jaderberg \bgroup \em et al.\egroup
  }{2016}]{e15}
Max Jaderberg, Karen Simonyan, Andrea Vedaldi, and Andrew Zisserman.
\newblock Reading text in the wild with convolutional neural networks.
\newblock {\em International Journal of Computer Vision}, 116(1):1--20, 2016.

\bibitem[\protect\citeauthoryear{Karatzas \bgroup \em et al.\egroup
  }{2013}]{f2}
Dimosthenis Karatzas, Faisal Shafait, Seiichi Uchida, Masakazu Iwamura,
  Lluis~Gomez i~Bigorda, Sergi~Robles Mestre, Joan Mas, David~Fernandez Mota,
  Jon~Almazan Almazan, and Lluis~Pere de~las Heras.
\newblock Icdar 2013 robust reading competition.
\newblock In {\em Proc. ICDAR}, pages 1484--1493, 2013.

\bibitem[\protect\citeauthoryear{Kingma and Ba}{2014}]{f4}
Diederik Kingma and Jimmy Ba.
\newblock Adam: A method for stochastic optimization.
\newblock {\em arXiv preprint arXiv:1412.6980}, 2014.

\bibitem[\protect\citeauthoryear{Lee and Osindero}{2016}]{e28}
Chen-Yu Lee and Simon Osindero.
\newblock Recursive recurrent nets with attention modeling for ocr in the wild.
\newblock In {\em Proc. CVPR}, pages 2231--2239, 2016.

\bibitem[\protect\citeauthoryear{Lucas \bgroup \em et al.\egroup }{2005}]{f1}
Simon~M Lucas, Alex Panaretos, Luis Sosa, Anthony Tang, Shirley Wong, Robert
  Young, Kazuki Ashida, Hiroki Nagai, Masayuki Okamoto, Hiroaki Yamamoto,
  et~al.
\newblock Icdar 2003 robust reading competitions: entries, results, and future
  directions.
\newblock {\em International journal on document analysis and recognition},
  7(2):105--122, 2005.

\bibitem[\protect\citeauthoryear{Mishra \bgroup \em et al.\egroup }{2012}]{e4}
Anand Mishra, Karteek Alahari, and CV~Jawahar.
\newblock Scene text recognition using higher order language priors.
\newblock In {\em Proc. BMVC}, 2012.

\bibitem[\protect\citeauthoryear{Novikova \bgroup \em et al.\egroup
  }{2012}]{e6}
Tatiana Novikova, Olga Barinova, Pushmeet Kohli, and Victor Lempitsky.
\newblock Large-lexicon attribute-consistent text recognition in natural
  images.
\newblock In {\em Proc. ECCV}, pages 752--765, 2012.

\bibitem[\protect\citeauthoryear{Rodriguez-Serrano \bgroup \em et al.\egroup
  }{2013}]{f8}
Jose~A Rodriguez-Serrano, Florent Perronnin, and France Meylan.
\newblock Label embedding for text recognition.
\newblock In {\em Proc. BMVC}, 2013.

\bibitem[\protect\citeauthoryear{Shi \bgroup \em et al.\egroup }{2013}]{e7}
Cunzhao Shi, Chunheng Wang, Baihua Xiao, Yang Zhang, Song Gao, and Zhong Zhang.
\newblock Scene text recognition using part-based tree-structured character
  detection.
\newblock In {\em Proc. CVPR}, pages 2961--2968, 2013.

\bibitem[\protect\citeauthoryear{Shi \bgroup \em et al.\egroup }{2016}]{e14}
Baoguang Shi, Xinggang Wang, Pengyuan Lyu, Cong Yao, and Xiang Bai.
\newblock Robust scene text recognition with automatic rectification.
\newblock In {\em Proc. CVPR}, pages 4168--4176, 2016.

\bibitem[\protect\citeauthoryear{Shi \bgroup \em et al.\egroup }{2017}]{e13}
Baoguang Shi, Xiang Bai, and Cong Yao.
\newblock An end-to-end trainable neural network for image-based sequence
  recognition and its application to scene text recognition.
\newblock {\em IEEE Trans. Pattern Analysis and Machine Intelligence},
  39(11):2298--2304, 2017.

\bibitem[\protect\citeauthoryear{Su and Lu}{2014}]{e12}
Bolan Su and Shijian Lu.
\newblock Accurate scene text recognition based on recurrent neural network.
\newblock In {\em Proc. ACCV}, pages 35--48, 2014.

\bibitem[\protect\citeauthoryear{Sutskever \bgroup \em et al.\egroup
  }{2014}]{e29}
Ilya Sutskever, Oriol Vinyals, and Quoc~V Le.
\newblock Sequence to sequence learning with neural networks.
\newblock In {\em Proc. NIPS}, pages 3104--3112, 2014.

\bibitem[\protect\citeauthoryear{Wang and Lu}{2017}]{f11}
Qingqing Wang and Yue Lu.
\newblock A sequence labeling convolutional network and its application to
  handwritten string recognition.
\newblock In {\em Proc. IJCAI}, pages 2950--2956, 2017.

\bibitem[\protect\citeauthoryear{Wang \bgroup \em et al.\egroup }{2011}]{e8}
Kai Wang, Boris Babenko, and Serge Belongie.
\newblock End-to-end scene text recognition.
\newblock In {\em Proc. ICCV}, pages 1457--1464, 2011.

\bibitem[\protect\citeauthoryear{Wang \bgroup \em et al.\egroup }{2012}]{e9}
Tao Wang, David~J Wu, Adam Coates, and Andrew~Y Ng.
\newblock End-to-end text recognition with convolutional neural networks.
\newblock In {\em Proc. ICPR}, pages 3304--3308, 2012.

\bibitem[\protect\citeauthoryear{Wikipedia}{2017}]{e18}
Wikipedia.
\newblock {Eye movement in reading}.
\newblock \url{https://en.wikipedia.org/wiki/Eye_movement_in_reading}, 2017.

\bibitem[\protect\citeauthoryear{Wu \bgroup \em et al.\egroup }{2017}]{e2}
Yi-Chao Wu, Fei Yin, and Cheng-Lin Liu.
\newblock Improving handwritten {C}hinese text recognition using neural network
  language models and convolutional neural network shape models.
\newblock {\em Pattern Recognition}, 65:251--264, 2017.

\bibitem[\protect\citeauthoryear{Yao \bgroup \em et al.\egroup }{2014}]{e11}
Cong Yao, Xiang Bai, Baoguang Shi, and Wenyu Liu.
\newblock Strokelets: A learned multi-scale representation for scene text
  recognition.
\newblock In {\em Proc. CVPR}, pages 4042--4049, 2014.

\bibitem[\protect\citeauthoryear{Ye and Doermann}{2015}]{e20}
Qixiang Ye and David Doermann.
\newblock Text detection and recognition in imagery: A survey.
\newblock {\em IEEE Trans. Pattern Analysis and Machine Intelligence},
  37(7):1480--1500, 2015.

\bibitem[\protect\citeauthoryear{Yin \bgroup \em et al.\egroup }{2017}]{e17}
Fei Yin, Yi-Chao Wu, Xu-Yao Zhang, and Cheng-Lin Liu.
\newblock Scene text recognition with sliding convolutional character models.
\newblock {\em arXiv preprint arXiv:1709.01727}, 2017.

\bibitem[\protect\citeauthoryear{Zhu \bgroup \em et al.\egroup }{2016}]{e35}
Yingying Zhu, Cong Yao, and Xiang Bai.
\newblock Scene text detection and recognition: Recent advances and future
  trends.
\newblock {\em Frontiers of Computer Science}, 10(1):19--36, 2016.

\end{thebibliography}

~\newpage
~\newpage

\appendix
%\titleformat{\section}{\centering\Huge\bfseries}{A }{1em}{}
\section{Supplementary Material of SCAN}

To give a more comprehensive understanding of SCAN (the abbreviation of Sliding Convolutional Attention Network), we further conduct the extended visualization experiments. First, we vividly give a step-by-step introduction of the recognition process in detail. Then, we give more visualization examples to demonstrate the superiority of SCAN.

\subsection{An Example of Recognition Process}
\label{recognition_example}

%??€????¡ã¡À??????????¡¤¡À?????????????¨C?????????????RNN?????????CTC ?¡¯?Attention?????????CNN?????¡ë?????¡°¨¨?????
%¨¨?????bottom-up top-down???¨¨?????
\begin{figure}[htb]
    \begin{center}
    \includegraphics[width=0.85\linewidth]{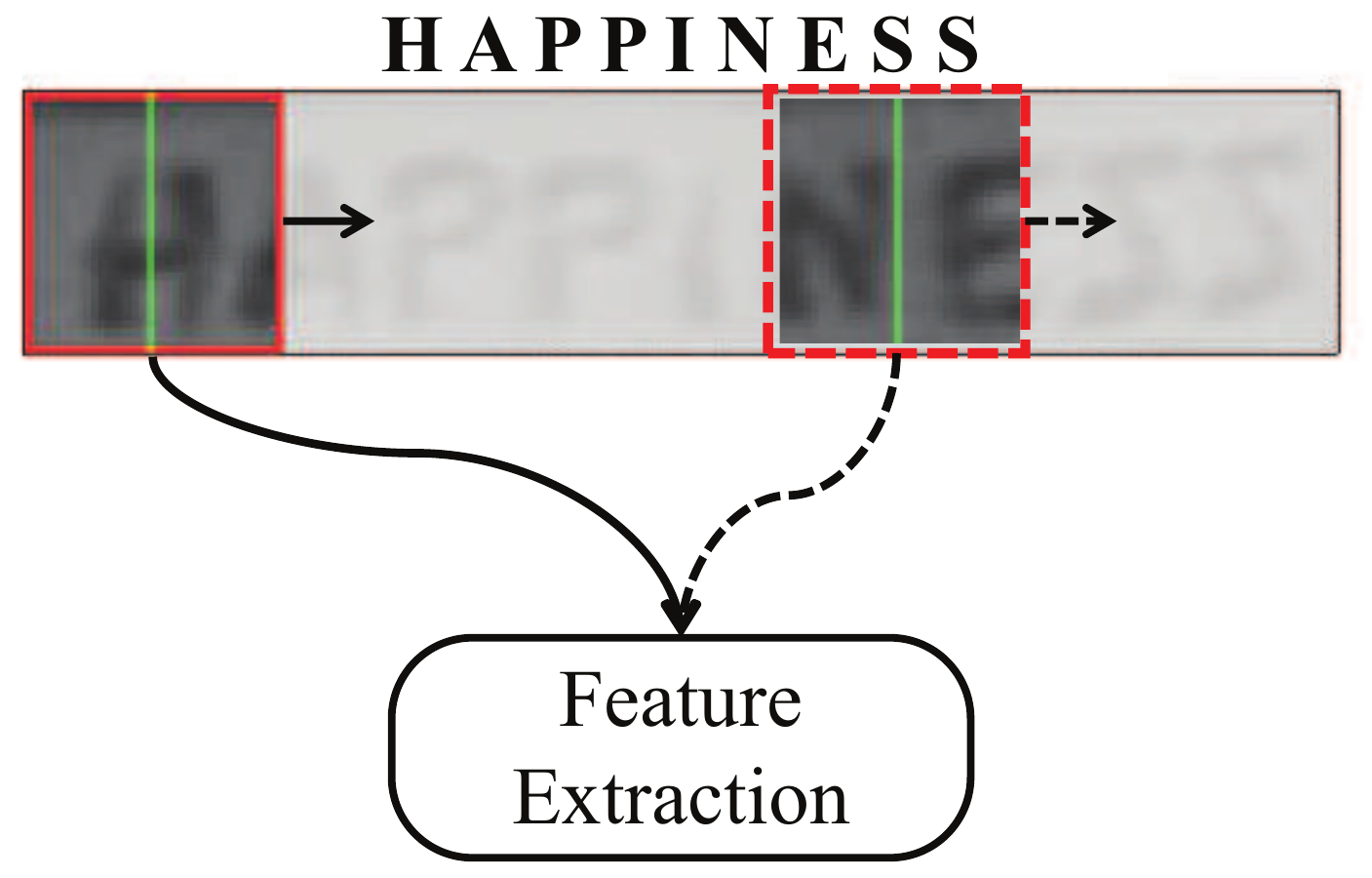}
    \end{center}
    \caption{Sliding window and feature extraction process for the textline image ``HAPPINESS".}
    \label{fig1s}
\end{figure}

\begin{figure}[phtb]
    \begin{center}
    \subfigure[The process of emitting the character ``A'']{
    \label{fig2sa}
    \includegraphics[width=0.4\textwidth]{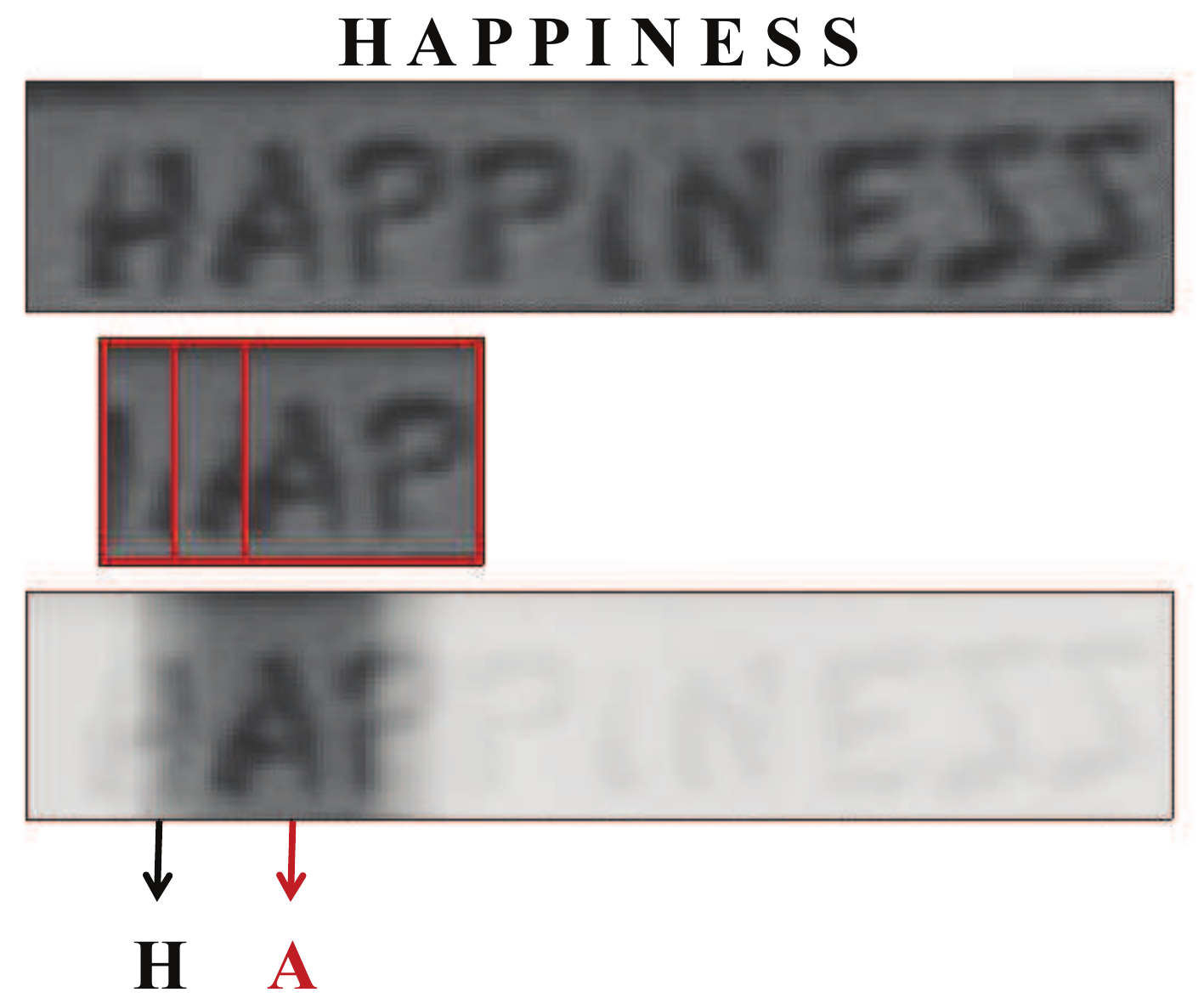}
    }

    \subfigure[The process of emitting the character ``N'']{
    \includegraphics[width=0.4\textwidth]{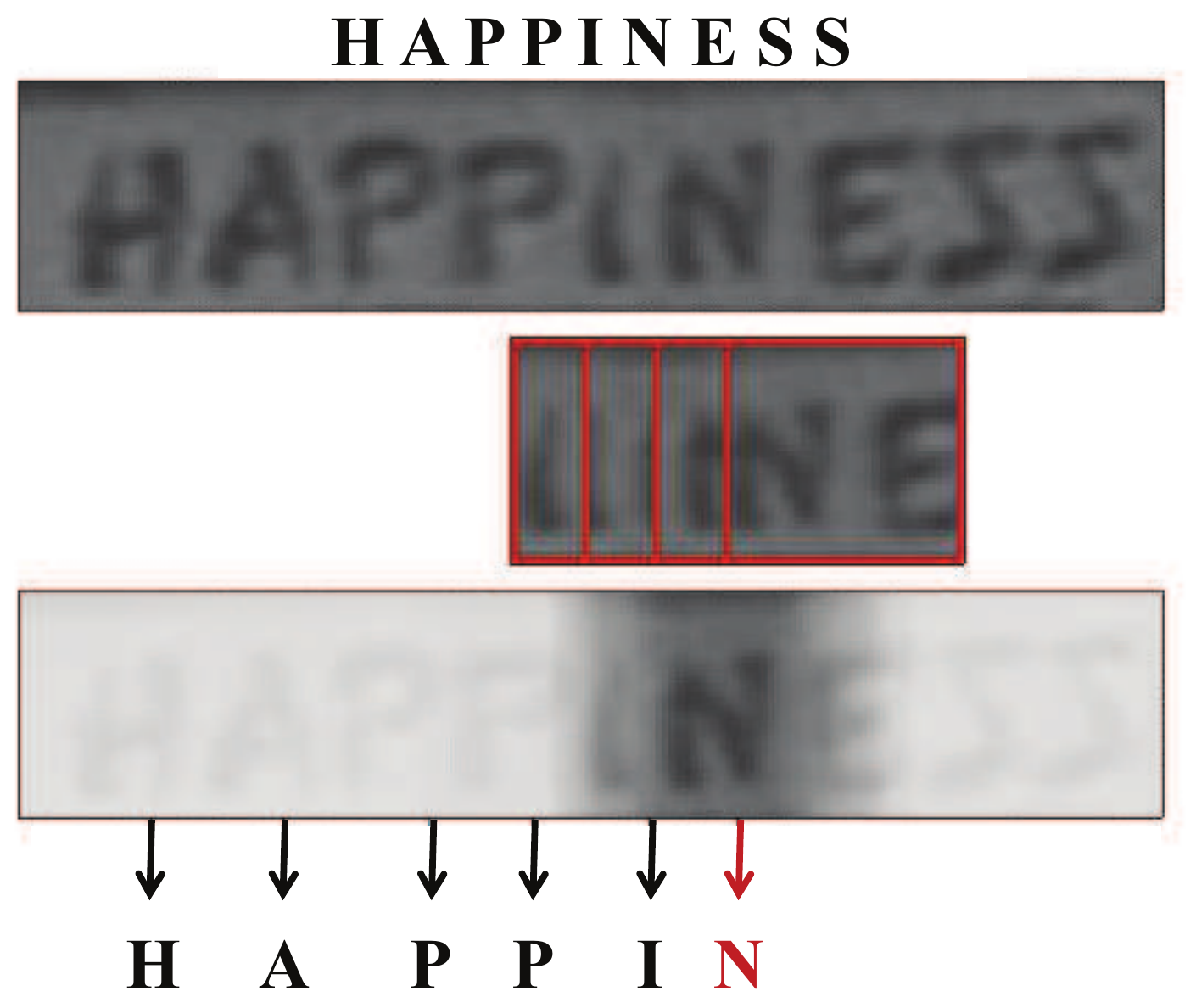}
    \label{fig2sb}
    }
    \end{center}
    \caption{The attention mechanism of SCAN based on the input feature sequence.}
    \label{fig2s}
\end{figure}

The SCAN recognition system consists of three major parts, namely a sliding window layer, a convolutional feature extractor, and a convolutional sequence network. Firstly, the sliding window layer splits the textline into overlapped windows. On the top of sliding window, a convolutional feature extractor is built to extract the discriminative features. These two steps are shown in Fig. \ref{fig1s} for the textline image ``HAPPINESS". Then, based on the extracted feature sequences, a convolutional sequence learning network is adopted to map the input to the output result. In Fig. \ref{fig2sa}, the character ``A'' is emitted according to the most relevant windows when considering the emitted previous characters and the whole feature sequence. Similar case is shown in Fig. \ref{fig2sb}, where the character ``N'' is emitted. The behavior of SCAN is very similar to the acuity of foveal vision in human reading.
The dynamic process of SCAN can be found in \url{https://github.com/nameful/SCAN}.

%------------------------------------------------------------------------
\subsection{Visualization of Recognition Results}
\label{visualization}

\begin{figure}[htb]
    \begin{center}
    \subfigure[Correct recognition samples]{
    \label{fig3sa}
    \includegraphics[width=0.45\textwidth]{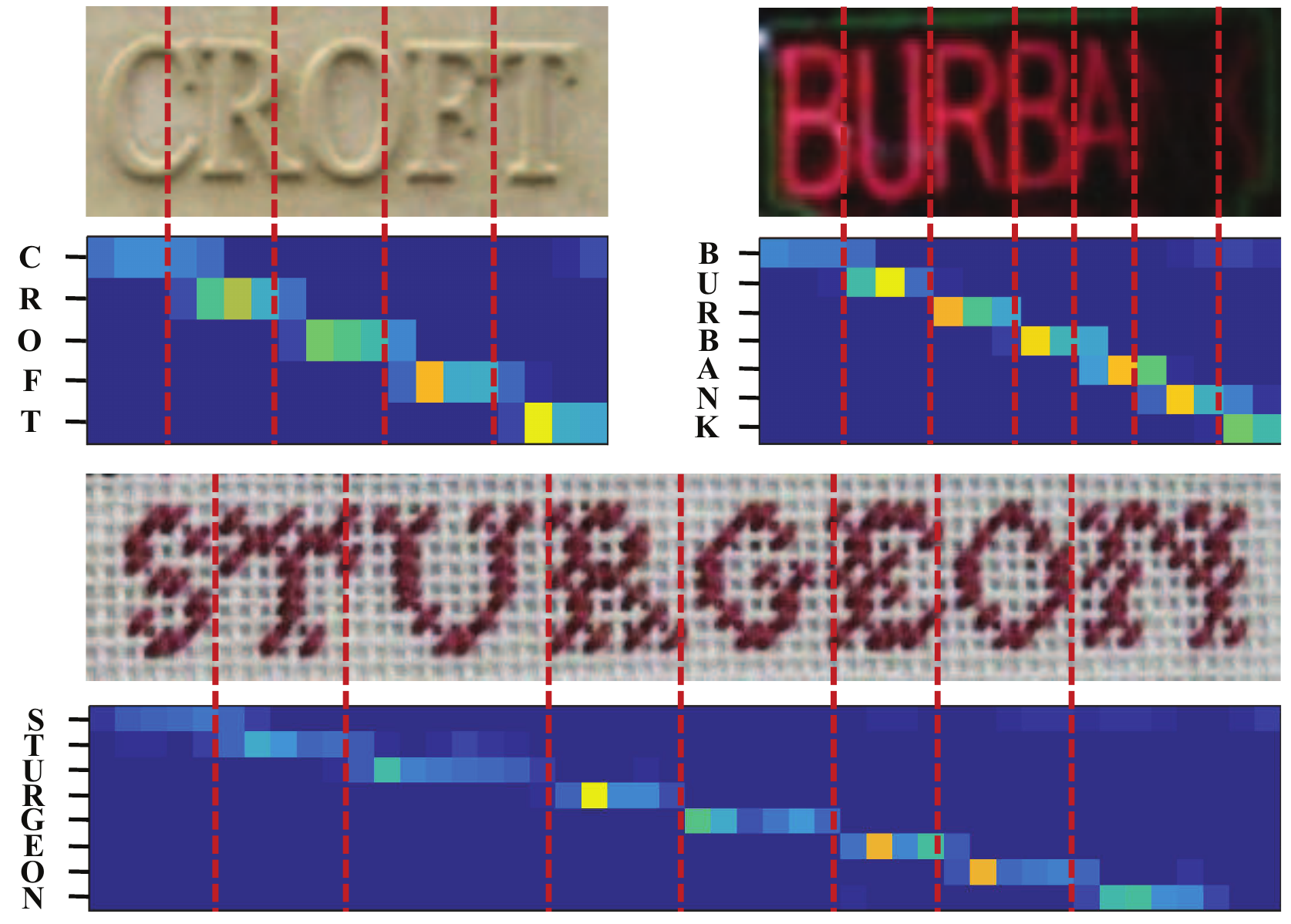}
    }

    \subfigure[Incorrect recognition samples. The ground truths of the three images are``\textbf{STARBUCKS}'',``\textbf{BOOKS}'' and ``\textbf{CENTRAL}'', respectively.]{
    \includegraphics[width=0.45\textwidth]{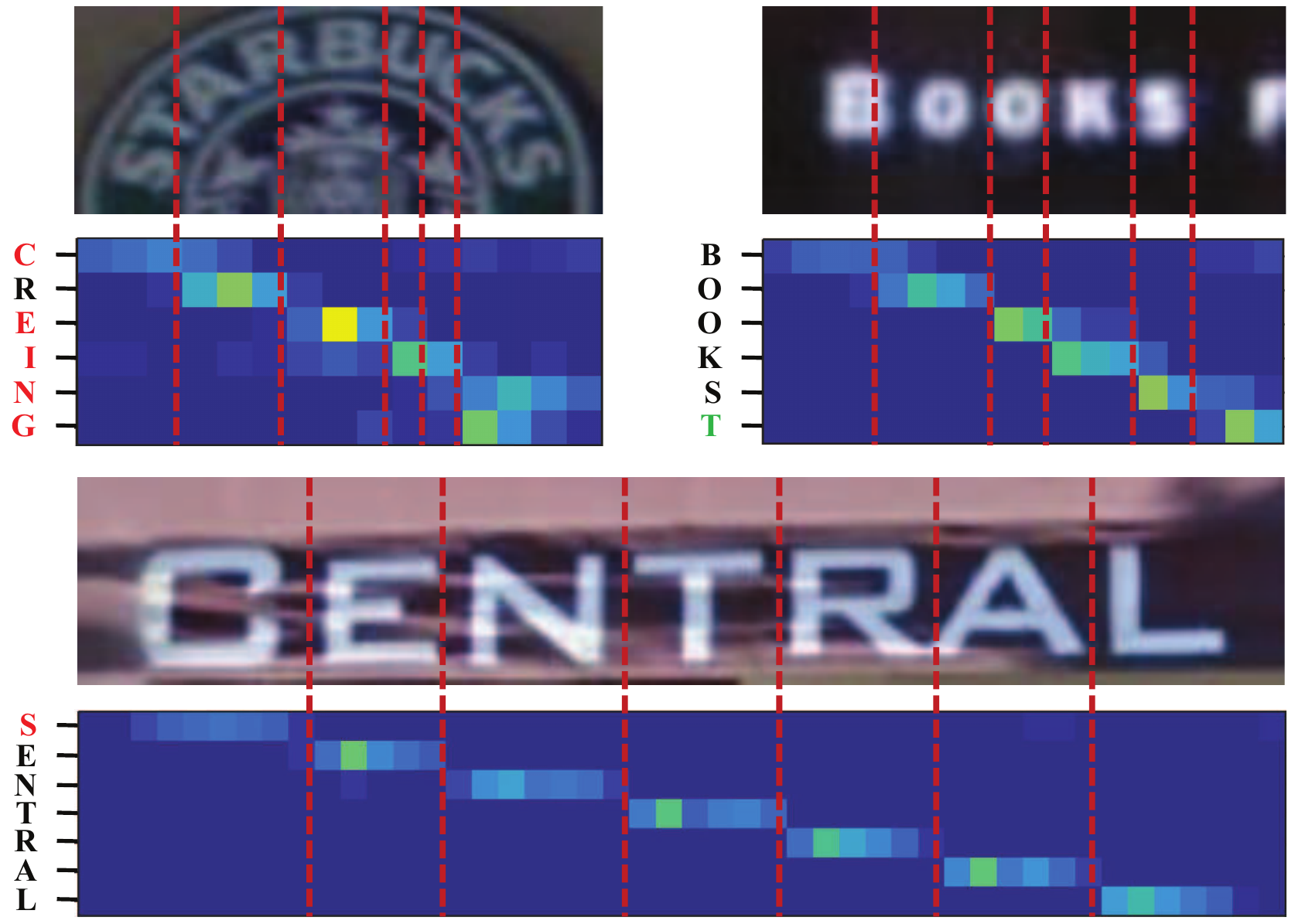}
    \label{fig3sb}
    }
    \end{center}
    \caption{Recognition examples of SCAN.}
    \label{fig3s}
\end{figure}

More visualization examples of the recognition results are shown in Fig. \ref{fig3s}, where each one is equipped with a corresponding attention heatmap. Fig. \ref{fig3sa} shows some correct recognition samples. We can see that SCAN can not only transcribe texts in slightly blurred or curved images robustly in most cases, but also locate the boundary of each character. Therefore, SCAN can accurately attend on the most relevant windows to give the final recognition results. We find that in some cases (such as the word image ``CROFT''), some attention weights far away from the specified character (``C'' in this case) also has certain responses in the heatmap. This phenomenon indicates that SCAN may have implicitly modeled the between-character relationship because of the long context information the convolutional model can capture. Moreover, we could make use of some heuristic rules to remove these weights if we need to locate the position of each character precisely, since they are usually restricted to a very small range with relatively low responses.

On the other hand, some incorrect recognition samples are shown in Fig. \ref{fig3sb}. It can be seen that SCAN are still unable to deal with severely blurred or curved word images, although it has powerful feature extractor and sequence learning network. The attention of SCAN is usually drifted in images containing insertion and deletion errors. It is an alternative approach to utilize better structure for feature extraction (such as DenseNet) to enhance the discriminant of features. For those irregular text images, we may first acquire the center curve of the text line by some detection\footnote{Nowadays, it is a trend to use direct regression for multi-oriented scene text detection.} or curve fitting techniques, and then slide along the curve to obtain the window sequence. In this way, we may further improve the performance of SCAN.

\end{document}